# Candidate Attended Dialogue State Tracking Using BERT

**Junyuan Zheng, Onkar Salvi, John Chan**
OneConnect US Research Institute, New York, NY, 10019
{zhengjunyuan616, onkar.salvi7, jchan.jc}@gmail.com

**Abstract**

Dialogue state tracking (DST) is one of the core components in task-oriented dialogue systems. At each turn in a conversation, DST estimates the user belief or dialogue state, which is used as input for downstream modules to predict system actions and generate responses. The increasingly popular dialogue system applications like Google Assistant, Siri and Alexa need to support a large number of services and APIs, resulting in growing attention to the scalability of such systems. Especially for some domains with little or no training data, the capability of transferring existing knowledge of other domains is highly desired. In this paper, we present a novel scalable framework for multi-domain dialogue state tracking. The proposed system leverages the pretrained BERT model to achieve zero-shot generalization, making it easy to quickly adapt to new domains without additional training. The performance of our model is evaluated on recently released schema-based dialogue (SGD) dataset, showing significant improvement compared to previous baseline.

## 1 Introduction

Task-oriented dialogue systems are playing an important role in facilitating our daily life in the form of personal virtual assistants, customer service agents, website chat systems, etc. These applications, by providing a conversational interface to different backends, help users easily accomplish tasks in a wide range of domains, such as restaurant reservation, flight booking, and music recommendation. While sequence-to-sequence learning (Sutskever, Vinyals, and Le 2014) has inspired lots of studies in building non-task-oriented dialogue systems, task-oriented systems still heavily rely on modularized pipeline. Typically, a task-oriented dialogue system has four components: a Natural Language Understanding (NLU) module semantically to parse the user utterance, a Dialogue State Tracking (DST) module to maintain the dialogue states throughout the conversation, a Dialogue Policy module to predict the next system action based on the current dialogue state, and a Natural Language Generation module to convert system actions into natural language responses (Rudnicky et al. 1999; Zue et al. 2000; Chen et al. 2017).

In recent years, the research community is paying significant attention to enhancing the scalability of dialogue system in order to quickly and easily add more functionalities. For popular commercial applications such as Google Assistant, Siri, or Alexa, the generalization capability is desired when developers want to integrate support for additional services or APIs. Given limited data in an unseen service or a new domain, the dialogue systems are expected to perform robustly without laborious data collection and annotation. Motivated by this expectation, several studies have focused on zero-shot language understanding and scalable dialogue state tracking (Bapna et al. 2017; Rastogi, Hakkani-Tür, and Heck 2017; Ren et al. 2018; Shah et al. 2019; Chao and Lane 2019).

Recently in the 8th Dialog System Technology Challenges (Seokhwan Kim 2019), Rastogi et al. (2019) introduce the Schema-Guided Dialogue (SGD) dataset, which exceeds existing task-oriented corpura like MultiWOZ (Budzianowski et al. 2018) in scale. In the proposed shcema-guided paradigm, a service's schema is defined by a set of intents and slots, where a natural language description is given for each intent and slot (as shown in Figure 1). Meanwhile, the emergence of pretrained models like BERT (Devlin et al. 2018) and XLNet (Yang et al. 2019) make it easier to utilize the natural language elements as inputs to achieve zero-shot generalization, since these models are pretrained on large corpura. Thanks to the advances in natural language processing, the path towards scalable dialogue system modeling becomes clearer.

In this work, we propose a dialogue system that is capable of dealing with a previously unseen domain via parameter sharing. The system focuses on NLU and DST components, both of which leverage BERT model to encode the natural language descriptions of intents and slots predefined in schema. In each dialogue turn, a slot tagger identifies non-categorical slot values mentioned in the user utterance, and a state tracker sequentially predicts the active intent, requested slots, and slot values of the user belief. Here, we frame DST as a natural language inference problem (Bowman et al. 2015). Given a user utterance as premise, we use a BERT model to identify the entailment relationships among a set of hypotheses. Varying from task to task, the hypothesis sentences could be: (1) intent descriptions for intent classification; (2) slot descriptions for requested slot detection; (3)

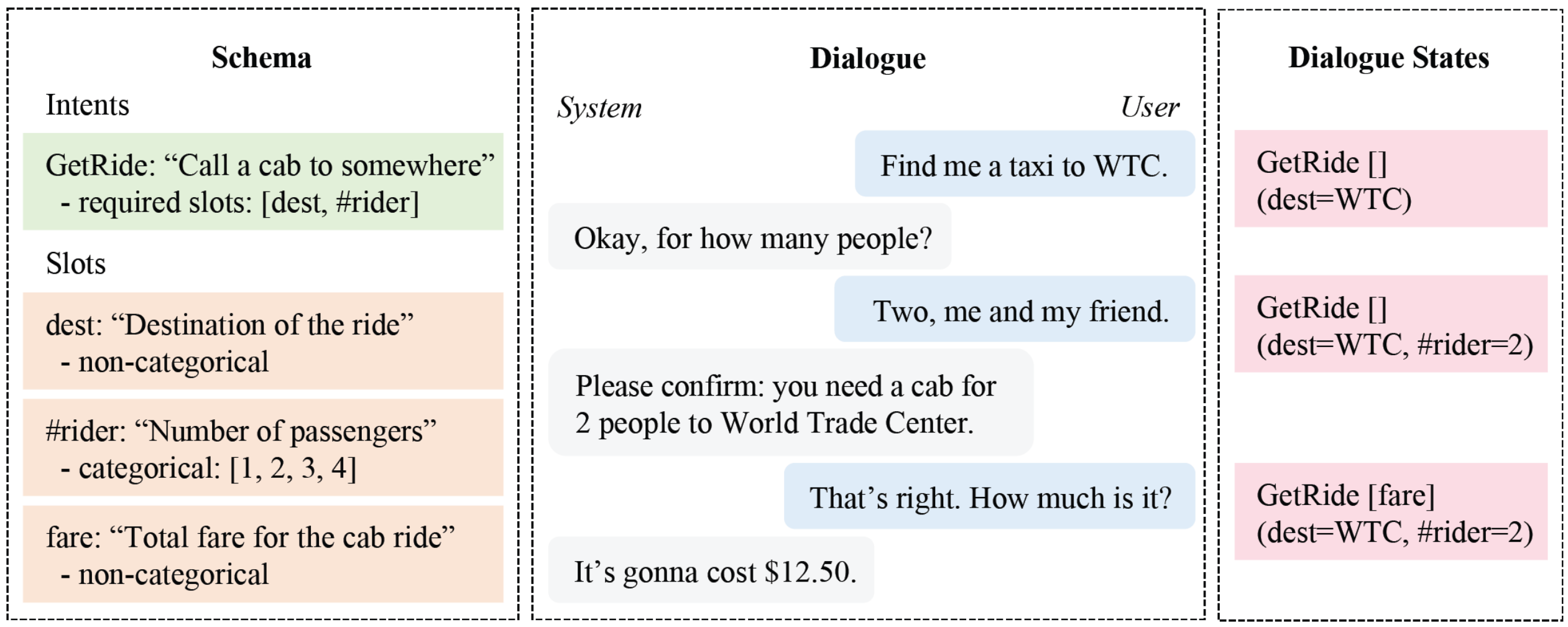

Figure 1: An example of dialogue for a taxi service. On the left demonstrates the schema of the service. For each intent and slot, a natural language description is given, along with other fields such as required slots for a intent or possible values for a categorical slot. The dialogue states of each user turn is displayed on the right panel, in the format of "$Intent$ [$requested_slots$] ($slot_values$)".

slot description and value pairs for candidate classification. By fine-tuning BERT weights, we expect the summary embedding (i.e. embedding of the [CLS] token) of the premise-hypothesis pair will capture enough information about the logic relationship, so as to facilitate DST (e.g. helping select the right candidate for a slot).

The rest of the paper is organized as follows: Section 2 describes related work. In Section 3, we detail our model architecture, followed by Section 4 presenting evaluation results and ablation studies. Finally, we discuss and conclude in Section 5 and 6.

## 2 Related Work

DST aims to maintain dialogue states in a complicated conversation. Dialogue state is usually represented as a set of slot-value pairs, which serves as parameters to interact with backend APIs and as inputs for downstream Dialogue Policy module to predict next system action. Traditional methods (Henderson, Thomson, and Young 2014; Mrkšić et al. 2015; Wen et al. 2016) predict the slot-value pairs by performing classification over a predefined candidate list. These models rely on delexicalization (i.e. replacing the slot types and values with generic tokens), which requires hand-crafted ontology with domain-specific lexicons. In practice, the ontology is not always available and the semantic dictionaries could be dynamically changing or hard to exhaust, limiting the generalization of these models. To this end, approaches have been proposed that score over a fixed (Mrkšić et al. 2016; Zhong, Xiong, and Socher 2018) or dynamic (Rastogi, Hakkani-Tür, and Heck 2017; Rastogi, Gupta, and Hakkani-Tur 2018; Ren et al. 2018) candidate list, extending the flexibility and scalablility of DST to rarely observed or unseen values in the training data. Other systems try to predict dialogue states directly from the dialogue context to handle previously unseen values, by applying pointer network (Xu and Hu 2018) or copy mechanism (Wu et al. 2019). These approaches are based on the assumption that the values are available in the conversation history, which may not be always true. To this end, some models combine the candidate-list-based and span-based approaches to take advantages from both sides (Goel, Paul, and Hakkani-Tür 2019; Zhang et al. 2019).

On the other hand, pretrained models such as BERT are gaining increasing attention in research community because of their promising performances on diverse downstream tasks. Many studies are seeking to leverage BERT in DST. Particularly, inspired by the success of BERT in reading comprehension tasks, most of the studies leveraging BERT in DST choose the span-based approach (Lee, Lee, and Kim 2019; Gao et al. 2019; Chao and Lane 2019; Zhang et al. 2019). In our study, we formulate DST as a natural language inference problem and utilize BERT to achieve zero-shot generalization.

## 3 Models

Our dialogue system consists of two major components: Natural Language Understanding (NLU) and Dialogue State Tracking (DST). These two components work successively to predict the user belief, i.e. dialogue state. Let each turn in a dialogue be a user utterance and its preceding system utterance. In the first stage, the slot tagger predicts all values of non-categorical slots mentioned in the user utterance. The tagged slot values are then used to update the candidate tracker. In the second stage, the user utterance is paired with intent descriptions, slot descriptions and slot-value pairs to get according embeddings from BERT. These embeddings are used as inputs of intent classifier, requested slot detector and candidate classifier respectively to get final predictions of the dialogue state.

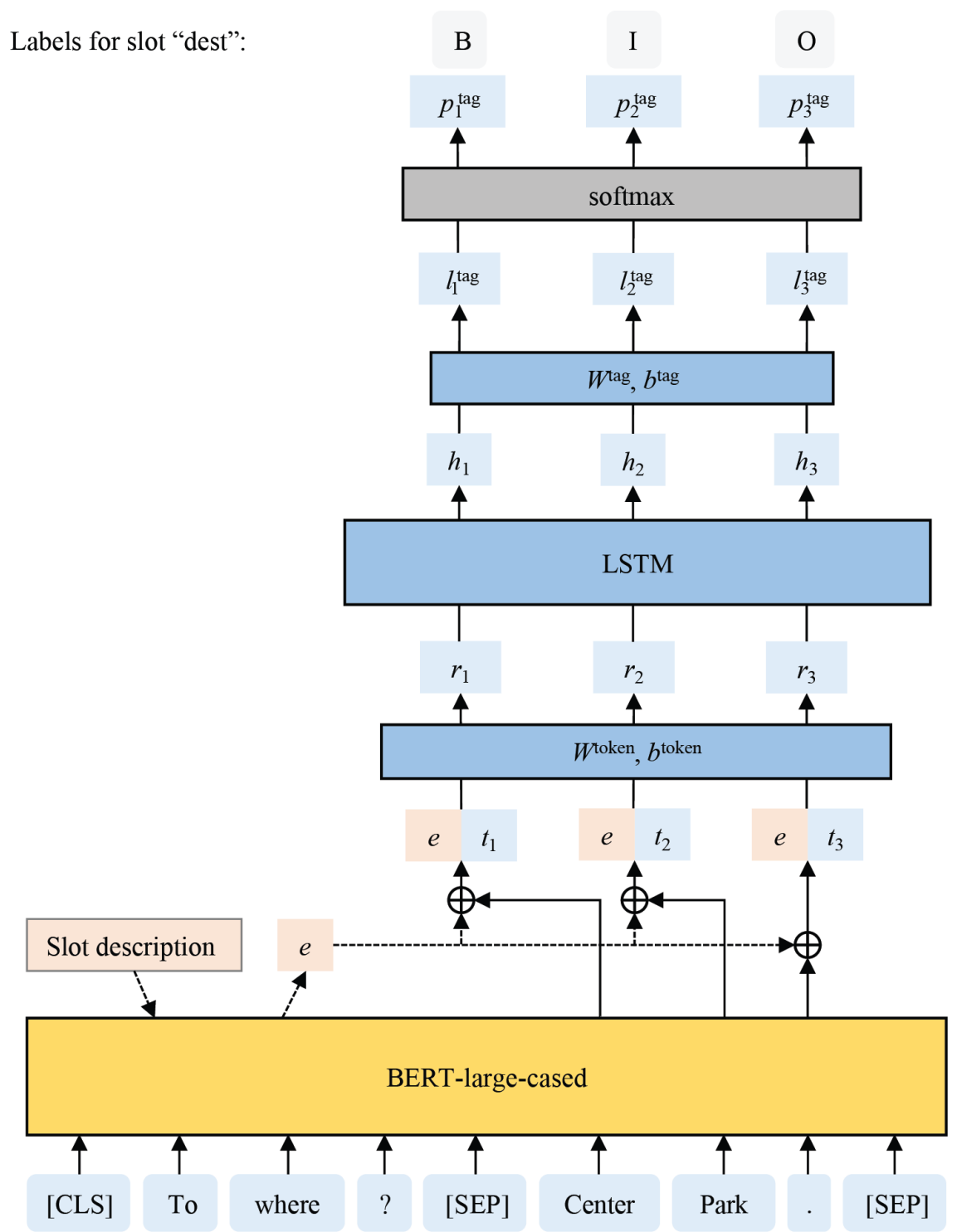

Figure 2: Slot tagger architecture.

## 3.1 Natural Language Understanding

**Slot Tagger** Slot tagger is the only model in the NLU component, whose task is to identify any possible values mentioned in the current user utterance for all non-categorical slots. We use IOB tagging schema introduced in (Sang and Buchholz 2000) to assign a label to each token $x_m$ in the user utterance (Figure 2), where $0 < m \leq M$, $M$ being the total number of tokens in the user utterance. The distribution over B, I and O tags is calculated using Equations 1–4. For a given non-categorical slot, the slot description is passed through BERT and the embedding of [CLS] token is captured as the summary embedding $e$. Then the system utterance and user utterance pair is encoded by BERT to obtain token level representations $t_m$. Each $t_m$ is concatenated with $e$ and then projected to get a representation $r_m$ through a linear layer. The sequence $\{r_m\}$ is fed into a bidirectional LSTM to get hidden states sequence $\{h_m\}$.

$$r_m = W^{\text{token}}(t_m \oplus e) + b^{\text{token}} \quad (1)$$

$$h_m = \text{LSTM}(r_m) \quad (2)$$

$$l_m^{\text{tag}} = W^{\text{tag}} h_m + b^{\text{tag}} \quad (3)$$

$$p_m^{\text{tag}} = \text{softmax}(l_m^{\text{tag}}) \quad (4)$$

Finally, $h_m$ is projected into a 3-dimension logit $l_m^{\text{tag}}$, and $p_m^{\text{tag}}$ is calculated using softmax. During inference, we predict the label for each token $x_m$ as $\text{argmax}(p_m^{\text{tag}})$, and values labeled with B and I tags are added into the tagged slots.

## 3.2 Dialogue State Tracking

**Intent Classifier** Active intent in a given service denotes a specific task the user wants to complete (e.g. finding a restaurant or booking a flight ticket). The user's active intent remains the same until s/he ends the conversation (i.e. "NONE" intent) or changes to another intent. In the middle of a dialogue, the user may talk about many things that are not strongly related to the active intent. In consideration of this, we predict the intent change instead of the actual active intent. More specifically, let $I$ be the intent set for a given service. $\phi$ and $\psi$ denote "NONE" and "UNCHANGED" intent respectively. The distribution over all intents for a given service is calculated as follows.

$$l_i^{\text{int}} = W^{\text{int}} r_i^{\text{int}} + b^{\text{int}} \quad (5)$$

$$l_\phi^{\text{int}} = W_\phi^{\text{int}} r^{\text{int}} + b_\phi^{\text{int}} \quad (6)$$

$$l_\psi^{\text{int}} = W_\psi^{\text{int}} r^{\text{int}} + b_\psi^{\text{int}} \quad (7)$$

$$p^{\text{int}} = \text{softmax}(l_\alpha^{\text{int}}),\ \alpha \in I \cup \{\phi, \psi\} \quad (8)$$

Here, $r_i^{\text{int}} = e_i \oplus a_i$ is the intent related feature where $e_i$ is the summary embedding of user utterance and intent description pair, and $a_i$ is a one-hot encoding indicating whether the specific intent is offered in preceding system acts, for intent $i \in I$. $r^{\text{int}} = u \oplus a$ where $u$ is the user utterance embedding and $a$ is a one-hot encoding indicating whether any intent is offered in preceding system acts. $W^{\text{int}}$, $b^{\text{int}}$, $W_\phi^{\text{int}}$, $b_\phi^{\text{int}}$, $W_\psi^{\text{int}}$, and $b_\psi^{\text{int}}$ are trainable parameters shared across different services. During inference, $\text{argmax}(p^{\text{int}})$ is taken as the intent label. If "UNCHANGED" intent $\psi$ is predicted, we use the same intent from the previous turn.

**Requested Slot Detector** Requested slots are the slots whose values the user is asking for (e.g. what is the address of the restaurant). Let $S$ be the slot set of a given service. The probability of a given slot $s \in S$ is calculated as below.

$$l_s^{\text{req}} = W^{\text{req}} e_s + b^{\text{req}} \quad (9)$$

$$p_s^{\text{req}} = \text{sigmoid}(l_s^{\text{req}}) \quad (10)$$

Here, $e_s$ is the summary embedding of user utterance and slot description pair for slot $s$. $W^{\text{req}}$, $b^{\text{req}}$ are trainable parameters. At inference time, we predict a slot as requested if the probability $p_s^{\text{req}} > 0.5$.

**Candidate Tracker** For each slot $s$ in a given service, we keep track of a candidate set $C_s$. A special candidate $\delta$ denoting "dontcare" is always in the set $C_s$. For categorical slots, the set $C_s = \{\delta\} \cup V_s$ will be constant, where $V_s$ is the set of possible values. For non-categorical slots, apart from candidate $\delta$, the most recent $K$ candidates are maintained in the set $C_s$, where $K$ is an arbitrary hyper-parameter. The candidate tracker is also used to calculate candidate related source features $a_{s,c}^{\text{src}}$.

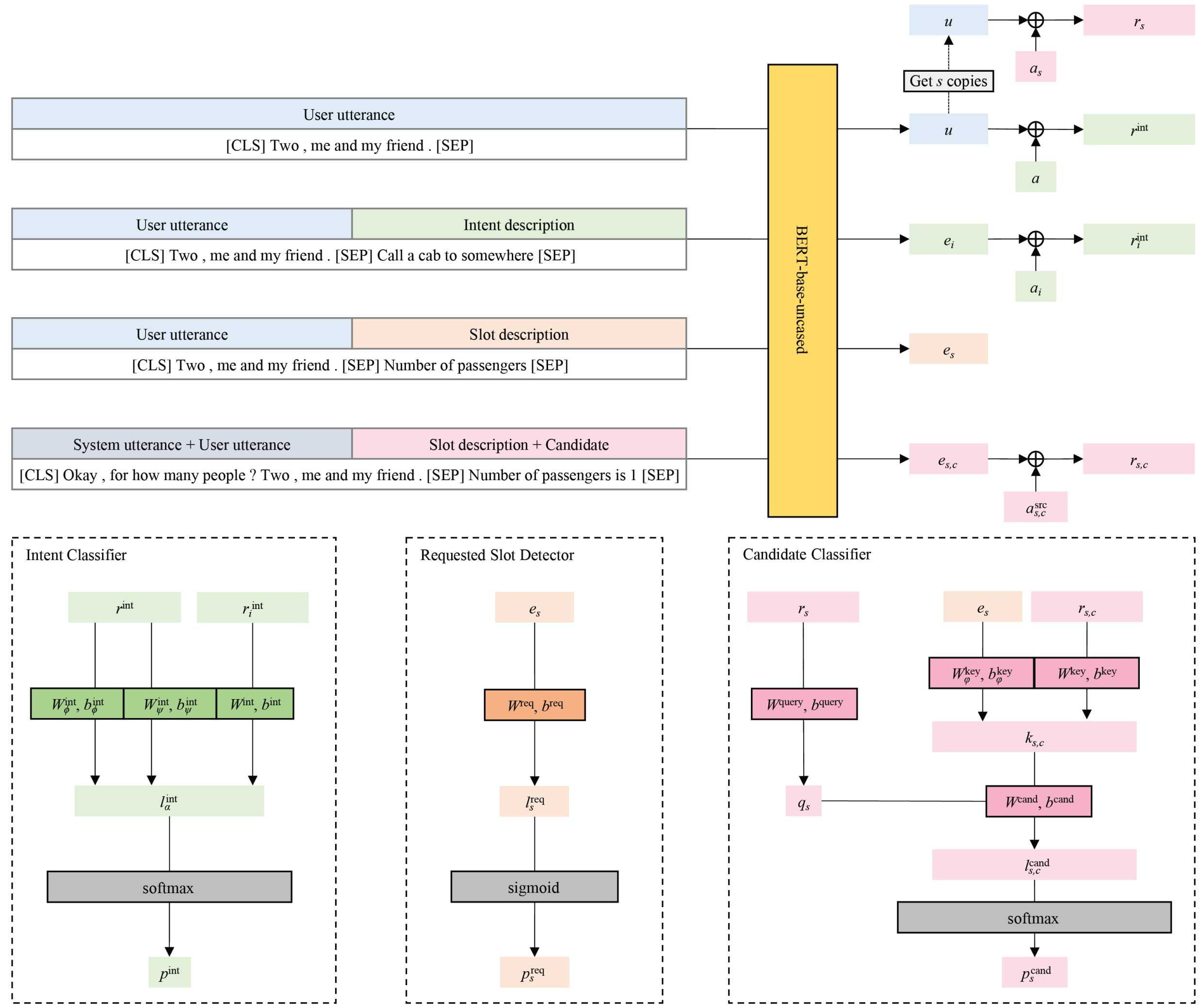

Figure 3: Architecture of Dialogue State Tracker.

**Candidate Classifier** Given a candidate set $C_s$ for slot $s$, the candidate classifier predicts the dialogue state updates by calculating the distribution over the candidate set. We define a dialogue state update for a slot $s$ as the difference between the dialogue state of current turn and that of preceding turn. If the system asks for confirmation and the user confirms the slot, the state is also considered updated, although the value in the current state may not be different from the preceding turn. When a slot $s$ is not updated, a special candidate label $\varphi$ is assigned, denoting the "none" state.

To predict the distribution over the candidate set, we use an architecture inspired by the attention mechanism (Graves, Wayne, and Danihelka 2014; Luong, Pham, and Manning 2015). First, for a given slot $s$, a slot related query vector $q_s$ is calculated by Equation 11. Here, $r_s = u \oplus a_s$, where $a_s$ is the one-hot encoding for slot related system acts. For each candidate $c$, the key vector is calculated by Equation 12. $r_{s,c} = e_{s,c} \oplus a_{s,c}^{\text{src}}$, where $e_{s,c}$ is the summary embedding of user utterance and slot-value pair. $W^{\text{query}}$, $b^{\text{query}}$, $W_{\varphi}^{\text{key}}$, $b_{\varphi}^{\text{key}}$, $W^{\text{key}}$, and $b^{\text{key}}$ are trainable parameter. Consequentially, slot related query vector $q_s$ attends over candidate related vectors $k_{s,c}$ via Equation 13 to get logits $l_{s,c}^{\text{cand}}$. Trainable parameters $W^{\text{cand}} \in \mathbb{R}^{d \times d}$, $b^{\text{cand}}$ here function similarly to general attention mechanism introduced in (Luong, Pham, and Manning 2015).

$$q_s = W^{\text{query}} r_s + b^{\text{query}} \tag{11}$$

$$k_{s,c} = \begin{cases} W_{\varphi}^{\text{key}} e_s + b_{\varphi}^{\text{key}} & \text{if } c = \varphi \\ W^{\text{key}} r_{s,c} + b^{\text{key}} & \text{for } c \in C_s \end{cases} \tag{12}$$

$$l_{s,c}^{\text{cand}} = q_s^{\text{T}} W^{\text{cand}} k_{s,c} + b^{\text{cand}} \tag{13}$$

$$p_s^{\text{cand}} = \text{softmax}(l_{s,c}^{\text{cand}}) \tag{14}$$

The logits $l_{s,c}^{\text{cand}}$ are normalized by softmax to get distribution over all candidate values. During inference time, the candidate with maximal probability is predicted. If a slot is predicted to be ”none”, its value is set the same as in preceding turn.

|  | **Active Int Acc** |  | **Req Slot F1** |  | **Average GA** |  | **Joint GA** |  | **Slot Tagging F1** |  |
|---|---|---|---|---|---|---|---|---|---|---|
|  | Dev | Test | Dev | Test | Dev | Test | Dev | Test | Dev | Test |
| SGD baseline | 0.908 | - | 0.973 | - | 0.740 | - | 0.411 | - | - | - |
| CA-DST | 0.945 | 0.923 | 0.989 | 0.982 | 0.949 | **0.895** | 0.835 | **0.695** | 0.977 | 0.959 |
| CA-DST+joint loss | 0.944 | 0.923 | **0.992** | **0.988** | **0.956** | 0.871 | **0.850** | 0.653 | 0.977 | 0.959 |
| (-pair encoding) | 0.943 | 0.912 | 0.973 | 0.961 | 0.884 | 0.840 | 0.674 | 0.578 | 0.977 | 0.959 |
| (-attention) | **0.951** | **0.932** | 0.991 | 0.986 | 0.882 | 0.783 | 0.673 | 0.468 | 0.977 | 0.959 |

Table 1: Evaluation results of our **C**andidate **A**ttended **D**ialogue **S**tate **T**racking (CA-DST) system. For Ablation study, all models are training with joint goal loss described in Section 4.1.

# 4 Evaluation

## 4.1 Experiments

We evaluate our system on SGD dataset. The sample dialogue and schema are described in Figure 1. In our system, NLU module and DST module are trained separately. For slot tagger, we use a large cased BERT model in a feature-based approach, without updating any of its parameters. For state tracker, we fine-tune a base uncased BERT model while the embedding layer and first 6 layers are fixed. This is a compromise due to the hardware limits. To facilitate convergence, we first fine-tune the BERT models on masked language modeling task using dialogues from SGD dataset before training each component. In DST module, intent classifier, requested slot detector and candidate classifier are trained jointly. Additionally, concerning all slots are treated independently, we add a joint goal loss penalizing incorrect prediction on what slots appear in the current turn. The joint goal loss is calculated as cross entropy of ”none” logits for all slots.

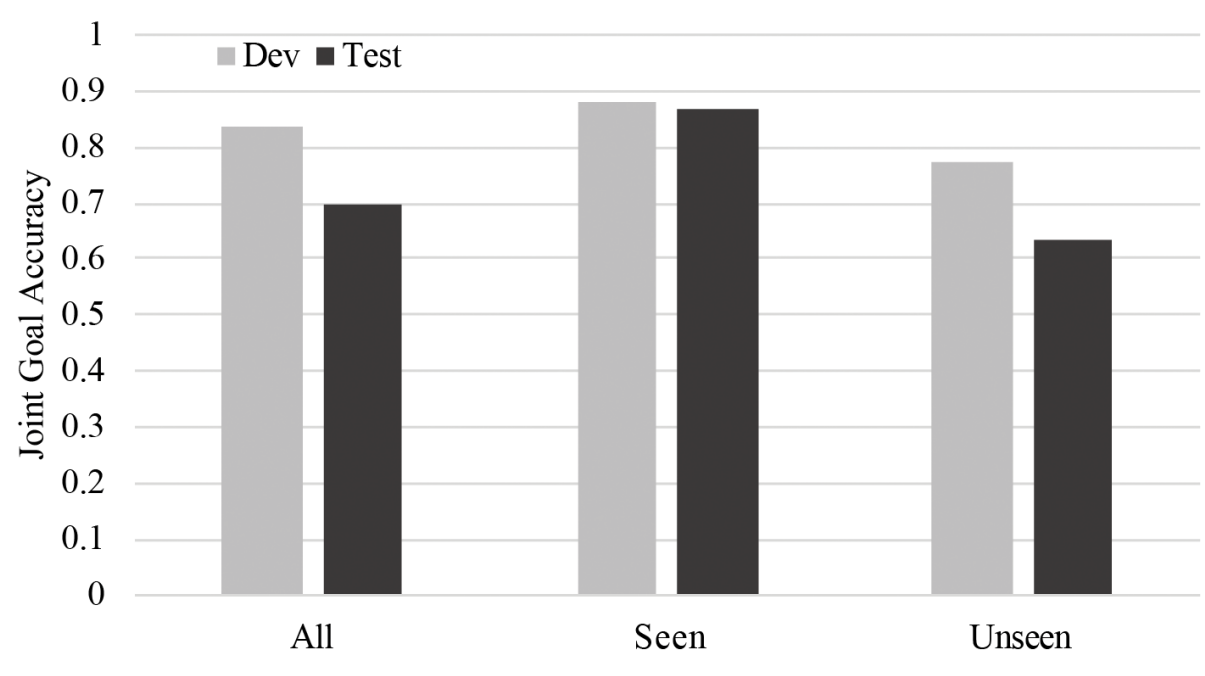

Figure 4: Joint Goal Accuracy on All services, services seen in training data, services not seen in training data.

We use the same metrics described in Rastogi et al. (2019) for evaluation: Active Intent Accuracy, Requested Slot F1, Average Goal Accuracy, Joint Goal Accuracy, Slot Tagging F1. The results are showed in Table 1. We also report here the results of the baseline model (Rastogi et al. 2019) as comparison.

As showed in Table 1, our system shows significant improvement in intent classification, requested slot detection and joint goal prediction compared to baseline, doubling the joint goal accuracy. However, the performance of our system exhibits a remarkable deterioration on the test set, indicating the models may suffer from over-fitting. Figure 4 shows the Joint Goal Accuracy in all, seen and unseen services. In those services that appear in training set, the model performs similarly, while in unseen services the Joint Goal Accuracy on test set is notably less than on dev set.

## 4.2 Ablation Studies

We perform ablation studies over two of the most important architecture designs of our system. The results are also shown in Table 1. The first one is that we pair each natural language element (intent description, slot description, or slot description plus value) with user utterance before entering BERT model. In this way the model can leverage the power of self-attention in BERT. To study the effectiveness of this pair encoding approach, we singly encode natural language elements and fuse the embeddings with user utterance embedding using the same projection layer described in baseline model (Rastogi et al. 2019). While in easier tasks like intent classification and requested slot prediction the performances are similar, the model without pair encoding shows a notable decline in Joint Goal Accuracy.

Second, we replace the attention-based bi-linear layer in the candidate classifier with feed forward layers. Model without attention performs significantly worse in Joint Goal Accuracy.

# 5 Discussion

The challenges towards scalable multi-domain DST come from three folds. First, during inference time, there are many unseen services which include several slots very different from those in training data. The only clues to identify those unseen slots are the natural language descriptions. Second, in multi-domain dialogues, the system is expected to infer certain slot values from other services when user changes a topic. Third, the expression diversity of natural language in dialogues further complicates the DST tasks. In this section we discuss the limitation and possible improvement direction based on these aspects.

In Table 2, we list Average Goal Accuracy and Joint Goal Accuracy per services in the dev set and test set. By examining those poorly performing services, we identify some chal-lenging cases. For example, in service "Movies_2" there is a slot "starring" with the description "Name of actor starring in movie". Similarly service "Movies_3" has a slot "cast" described as "Actors in the movie". Both unseen services include two slots of the person name type: one for director and one for actor. However, in training set, there is only one slot for director slot, which demands for higher level of generalization. The system not only needs to understand the actor slots refer to person names, but also has to understand that the person should be the actor instead of the director. This requires building a semantic correlation between word "actor" and others such as "act", "starring" or "plays". The BERT pretrained on large-scale corpura should have some capability to resolve such semantic ambiguity, but it seems to fail. Moreover, some cases are even difficult for human. For example, in the utterance "Do you have any movies with Mike Colter?", one needs to know that Mike Colter is an actor to correctly identify this value as an actor instead of a director. Possibly, the system may leverage external knowledge to deal with such problems.

| Service | Average GA | | Joint GA | |
|---|---|---|---|---|
| | Dev | Test | Dev | Test |
| Alarm_1* | 0.929 | 0.939 | 0.927 | 0.909 |
| Banks_2* | 0.982 | - | 0.952 | - |
| Buses_1 | 0.973 | - | 0.919 | - |
| Buses_3* | - | 0.896 | - | 0.622 |
| Events_1 | 0.981 | - | 0.913 | - |
| Events_3* | - | 0.895 | - | 0.681 |
| Flights_3* | 0.836 | - | 0.470 | - |
| Flights_4* | - | 0.681 | - | 0.322 |
| Homes_1 | 0.987 | - | 0.938 | - |
| Homes_2* | - | 0.956 | - | 0.563 |
| Hotels_1 | 0.949 | - | 0.884 | - |
| Hotels_2 | - | 0.928 | - | 0.859 |
| Hotels_4* | 0.936 | 0.939 | 0.844 | 0.859 |
| Media_2* | 0.911 | - | 0.761 | - |
| Media_3* | - | 0.791 | - | 0.594 |
| Messaging_1* | - | 0.769 | - | 0.683 |
| Movie_1 | - | 0.957 | - | 0.819 |
| Movie_2* | 0.911 | - | 0.761 | - |
| Media_3* | - | 0.683 | - | 0.700 |
| Music_1 | 0.936 | - | 0.823 | - |
| Music_3* | - | 0.754 | - | 0.299 |
| Payment_1* | - | 0.846 | - | 0.608 |
| RentalCars_1 | - | 0.947 | - | 0.772 |
| RentalCars_3* | - | 0.897 | - | 0.486 |
| Restaurants_2* | 0.947 | 0.949 | 0.747 | 0.730 |
| RideSharing_1 | 0.958 | - | 0.909 | - |
| RideSharing_2 | - | 0.944 | - | 0.840 |
| Services_1 | - | 0.987 | - | 0.960 |
| Services_4* | 0.989 | 0.986 | 0.892 | 0.876 |
| Trains_1* | - | 0.895 | - | 0.666 |
| Travel_1 | 0.965 | 0.862 | 0.930 | 0.778 |
| Weather_1* | 0.968 | 0.966 | 0.929 | 0.916 |

Table 2: Model performance per service. Services masked with "*" denote that they are unseen in the training set.

Another challenge is the slot transfer in multi-domain dialogues. For instance, when a user just checked the weather in Toronto and asks to book a hotel, the system is expected to assume the city slot in hotel domain is Toronto. Because there are unseen domains, it is difficult to come up with an exhausted slot-to-slot mapping list across different domains. Currently, our system collects all possible slot values from previous domain as the candidates, and relies on the semantic validity provided by BERT to solve the slot transferring problem (e.g. the probability of "The location of the hotel is Toronto" is much higher than that of "The location of the hotel is today"). This approach is not robust enough especially in certain domains like Flights, where it is hard to tell a city appeared in the dialogue history would be transferred to destination or origin. Additionally, in a more complicated situation, this slot-to-slot mapping could be dynamic, depending on the dialogue history. Take Flights and Hotels as example. A user just booked a flight from NY to LA, and then asks for a hotel on Friday. Practically, if the flight date is Thursday, we assume the user is looking for a hotel in LA. Yet, if the flight starts on Saturday, one should expect the user wants the hotel to be in NY. To distinguish such nuances, a dialogue system needs to incorporate adequate level of common sense reasoning.

Finally, the diversified expressions in natural language sometimes make it hard to predict dialogue states. For categorical slots, possible slot values are given in advance, where the task of DST is to map the words or phrases in user utterance to one of the candidate values. In most cases the mapping is trivial (e.g. "three" to "3", "checking" to "checking account"). However, sometimes the mapping could be logically difficult for machine if not for human. For example, "two years ago" for categorical slot "year" is mapped to "2017" given this year is 2019, which involves elementary level of mathematical calculation. This mapping may even change as time goes. As for non-categorical slots, the Joint Goal Accuracy is highly depending on the outputs of slot tagger. One major drawback of separating slot tagger and dialogue state tracker is the error propagation. If a slot is failed to be tagged by the NLU module, the downstream DST component would not select the correct candidate. Nevertheless, the pipeline approach on the other hand make it more flexible to plug in additional modules. By incorporating open-domain Named Entity Resolution (NER) tools or domain-specific slot tagger, the system is easy to fix some poorly performed domains without significant architecture changes.

## 6 Conclusions

To address the scalability challenge in task-oriented dialogue state tracking, we frame the problem as a natural language inference problem and propose a multi-domain schema-based dialog system. The system leverages pre-trained BERT model to achieve zero-shot generalization via parameter sharing. The experiments on the newly released Schema-Guided Dialogue dataset show the effectiveness of the proposed system.

## Acknowledgement

We would like to thank Surya Kasturi, Xin Chen, and anonymous reviewers for their constructive discussions and feedback.